\pdfoutput=1
\documentclass{article}
\usepackage[margin=1in]{geometry}
\usepackage{natbib}

\usepackage{booktabs}
\usepackage{float}
\usepackage{graphicx}
\usepackage{amsmath,amssymb,mathtools}
\usepackage{microtype}
\usepackage{xcolor}
\usepackage{enumitem}
\setlist[itemize]{leftmargin=1.3em,itemsep=1pt,topsep=2pt,parsep=0pt}
\setlist[enumerate]{leftmargin=1.3em,itemsep=1pt,topsep=2pt,parsep=0pt}
\usepackage{multirow}
\usepackage{array}
\usepackage{makecell}
\usepackage{subcaption}
\usepackage{placeins}
\usepackage{needspace}

\usepackage{tikz}
\usetikzlibrary{arrows.meta,positioning,shapes.geometric}
\usepackage{url}
\usepackage{hyperref}
\hypersetup{colorlinks=true,citecolor=blue,linkcolor=blue,urlcolor=blue}
\graphicspath{{figures/}}

\newcolumntype{P}[1]{>{\raggedright\arraybackslash}p{#1}}

\newcommand{\bbcell}[2]{%
  \parbox[t]{\linewidth}{\raggedright
  \textbf{top:} #1\\
  \textbf{vs.:} #2}%
}
\newcommand{\bbtie}[2]{%
  \parbox[t]{\linewidth}{\raggedright
  \textbf{top:} #1\\
  \textbf{tie:} #2}%
}
\newcommand{\bbnote}[2]{%
  \parbox[t]{\linewidth}{\raggedright
  \textbf{top:} #1\\
  \textbf{note:} #2}%
}

\title{Operational Feature Fingerprints of Graph Datasets via a White-Box Signal-Subspace Probe}

\author{%
\begin{minipage}{\textwidth}
\begin{tabular*}{\textwidth}{@{\extracolsep{\fill}} l r}
\begin{minipage}[t]{0.68\textwidth}
\vspace{0pt}
{\large\textbf{Xiong Yuchen}}\\
\textit{China-ASEAN College of Marine Sciences}\\
\textit{Xiamen University Malaysia}\\
\textit{Sepang 43900, Selangor, Malaysia}\\[0.8em]
{\large\textbf{Yeap Swee Keong}}\textsuperscript{*}\\
\textit{China-ASEAN College of Marine Sciences}\\
\textit{Xiamen University Malaysia}\\
\textit{Sepang 43900, Selangor, Malaysia}\\[0.8em]
{\large\textbf{Ban Zhen Hong}}\textsuperscript{*}\\
\textit{School of Energy and Chemical Engineering}\\
\textit{Xiamen University Malaysia}\\
\textit{Sepang 43900, Selangor, Malaysia}
\end{minipage}
&
\begin{minipage}[t]{0.28\textwidth}
\vspace{0pt}
\raggedleft
\texttt{yuchenak05@gmail.com}\\[5.6em]
\texttt{skyeap@xmu.edu.my}\\[5.6em]
\texttt{bzhong@xmu.edu.my}
\end{minipage}
\end{tabular*}\\[1.2em]
\centering\textsuperscript{*}Corresponding authors.
\end{minipage}%
}

\date{}

\newcommand{\Prw}{P_{\mathrm{row}}}
\newcommand{\Psy}{P_{\mathrm{sym}}}
\newcommand{\Ffull}{F^{(0)}}
\newcommand{\Fsel}{F}
\newcommand{\Ftr}{F_{\mathrm{tr}}}
\newcommand{\Ytr}{Y}
\newcommand{\Train}{\mathcal{T}}
\newcommand{\Sel}{\mathcal{S}}
\newcommand{\Blocks}{\mathcal{B}}
\newcommand{\BlockIdx}{\mathcal{I}}
\newcommand{\Alphas}{\mathcal{A}}
\newcommand{\Eval}{\mathcal{U}}
\newcommand{\Rraw}{R_{\mathrm{raw}}}
\newcommand{\Rlow}{R_{\mathrm{low}}}
\newcommand{\Rhigh}{R_{\mathrm{high}}}
\newcommand{\SRC}{\textsc{WG-SRC}}
\newcommand{\dataset}[1]{\textsc{#1}}

\begin{document}

\maketitle

\begin{abstract}
Graph neural networks achieve strong node-classification performance, but learned
message passing entangles ego features, neighborhood smoothing, high-pass graph
differences, class geometry, and classifier-boundary effects inside opaque
representations. This makes it difficult to determine why nodes are classified
as they are, and which graph-learning mechanisms are useful, harmful, or
necessary for a given dataset.

We propose \SRC{} (\emph{White-box Graph Signal--Subspace Residual Classifier}),
a white-box signal-subspace probe for prediction and graph dataset diagnosis.
\SRC{} replaces learned message passing with an explicit, named graph-signal
dictionary containing raw features, row- and symmetric-normalized low-pass
propagation, and high-pass graph differences. It then combines Fisher
coordinate selection, class-wise PCA subspaces, closed-form multi-$\alpha$
ridge classification, and validation-based score fusion. Because every signal
block and decision module is explicit, the fitted scaffold produces both
predictions and an operational fingerprint over raw-feature, low-pass,
high-pass, class-geometric, and ridge-boundary mechanisms.

Across six node-classification datasets, \SRC{} remains competitive with
aligned reproduced baselines and achieves a positive average gain under matched
repeated splits. Its fingerprints distinguish low-pass-dominated Amazon graphs,
mixed high-pass and class-geometrically complex Chameleon behavior, and raw- or
boundary-sensitive WebKB graphs. Aligned interventions further show that these
fingerprints are operational: they identify when high-pass blocks behave like
removable noise, when graph-derived or raw signals should be preserved, and
when ridge-type boundary correction matters. Additional fixed black-box
component probes further show that measured dataset fingerprints organize
architectural behavior across multiple black-box families: different measured
dataset conditions repeatedly favor different inductive biases. 
Thus, \SRC{} serves both as a functioning white-box classifier and as a dataset-fingerprinting probe, enabling fingerprint-conditioned analysis of how black-box graph-model components behave under different measured dataset conditions.
\end{abstract}

\section{Introduction}
Graph neural networks learn by repeatedly aggregating features over edges and applying trained transformations. This recipe is powerful, but it hides several mechanisms behind parameters: whether a prediction is driven by ego attributes, one-hop smoothing, two-hop return structure, high-pass differences between a node and its neighborhood, or a final discriminative boundary is often unclear. This opacity is especially problematic on heterophilic or mixed-homophily graphs, where naive smoothing can hurt because neighbors need not share labels \citep{pei2020geomgcn,zhu2020beyond,lim2021linkx}.

This paper starts from a different premise. Instead of learning a hidden
message-passing representation, we explicitly build the graph signals that a
shallow graph model might exploit, and classify them using linear-algebraic
modules whose behavior is inspectable. The resulting method, \SRC{} (\emph{White-box Graph Signal--Subspace Residual Classifier}),
constructs a named graph signal dictionary, selects discriminative coordinates, fits
class-wise PCA subspaces, fits a closed-form ridge boundary, and fuses the two
scores by validation. In this sense, the method is deliberately built around
subspace geometry and controlled low-rank structure: the same explicit
dimension-reduced objects are used both to make decisions and to analyze which
graph mechanisms a dataset is using.

The central point is that \SRC{} is a graph-dataset probe implemented through a
functioning white-box predictor. After validation selection, the fitted
scaffold provides both predictions and a mechanism readout: because every
signal block and decision module is named, the same fitted variables define
raw-feature reliance, low-pass propagation reliance, high-pass sensitivity,
class-subspace complexity, and ridge-boundary dependence. We use
\emph{white-box scaffold} for the fitted predictive model and
\emph{dataset probe} for the same scaffold when it is used as a measurement
instrument.

This view leads to three empirical requirements. First, the scaffold must be
predictively functional, so that its readout is produced by a model that has
captured useful dataset structure. Second, the fingerprint should be
operational: datasets assigned to different signal or decision regimes should
respond differently to aligned white-box interventions. Third, the measured
dataset condition should organize behavior beyond the \SRC{} scaffold itself,
as tested by fixed black-box component probes. The main text follows this
chain through predictive validity, mechanistic interventions, dataset
fingerprints, and fingerprint-conditioned black-box probing.

Appendix~\ref{app:paired-stability} reports paired random-split stability,
Appendix~\ref{app:atlas-figures} provides additional atlas views, and
Appendix~\ref{sec:blackbox-component-probing} gives the full black-box
component tables.

\medskip
\noindent\textbf{Contributions.} We make three contributions.
\begin{enumerate}[leftmargin=1.3em,itemsep=1pt,topsep=2pt]
    \item We introduce \SRC, a white-box graph classifier that replaces learned
    message passing with an explicit multi-hop graph-signal dictionary and
    replaces hidden representation layers with class-wise PCA subspaces,
    closed-form multi-alpha ridge regression, and validation-based score
    fusion.

    \item We show that the fitted variables of the same classifier induce an
    operational graph-dataset fingerprint. The fingerprint aggregates
    node-level raw-feature, low-pass, high-pass, class-geometric, and
    boundary-based readouts into dataset-level summaries of signal composition,
    class-subspace complexity, PCA--Ridge decision structure, and
    correct-versus-wrong signal shifts.

    \item We validate the fingerprint as a diagnostic object. \SRC{} remains
    competitive under aligned reproduced benchmarks, and its measured
    fingerprints agree directionally with white-box interventions,
    signal/error shifts, PCA--Ridge phase structure, and
    fingerprint-conditioned black-box component probes. These results connect
    dataset fingerprints to post-evaluation guidance for suppressing noisy
    high-pass blocks, preserving useful graph-derived or raw signals,
    strengthening boundary decisions, or improving class-specific subspace
    modeling.
\end{enumerate}

\section{Related Work}
\textbf{Graph neural networks and heterophily.} GCN \citep{kipf2017gcn}, GraphSAGE \citep{hamilton2017graphsage}, and GAT \citep{velickovic2018gat} learn node representations by aggregating local neighborhoods. Heterophilic graphs expose the limitations of pure smoothing. Geom-GCN \citep{pei2020geomgcn}, H2GCN-style designs \citep{zhu2020beyond}, adaptive PageRank filters \citep{chien2021gprgnn}, and LINKX \citep{lim2021linkx} address non-homophily by changing propagation, decoupling ego and neighborhood features, or using strong simple baselines. \SRC{} follows the decoupling intuition but removes learned hidden layers: it constructs low-pass and high-pass signals explicitly and audits which ones are used.

\textbf{White-box and subspace learning.} PCA and ridge regression are classical tools with transparent objectives \citep{pearson1901pca,hoerl1970ridge}. MCR$^2$ and ReduNet provide a modern white-box perspective on representation learning: classes should occupy structured, discriminative subspaces, and networks can be derived from optimization principles rather than treated as opaque stacks \citep{yu2020mcr2,chan2022redunet,wang2024mcr2landscape}. \SRC{} borrows the subspace viewpoint, but adapts it to graphs by first decomposing
node features into named graph signal blocks and then fitting class subspaces in
that signal space. Its white-box character therefore comes not only from using
explicit graph filters, but also from using subspace geometry, low-rank energy
control, and closed-form decision modules as analyzable mathematical objects.

\textbf{Transformer-style, latent-graph, and adaptive black-box graph models.}
Recent black-box graph learners explore several architectural directions
beyond classical local aggregation, including simplified global token mixing in
SGFormer \citep{wu2023sgformer}, global attention in GOAT
\citep{kong2023goat}, learned latent graph structure in NodeFormer
\citep{wu2022nodeformer}, and adaptive expert routing in GNNMoE
\citep{chen2024gnnmoe}. These models can be strong, but their gains are often
difficult to attribute to specific graph-learning mechanisms. We therefore use
selected black-box families and fixed component ablations as external probes:
after \SRC{} measures a dataset fingerprint, these ablations test whether the
measured raw-feature, low-pass, high-pass, class-geometric, or
boundary-sensitive condition favors corresponding architectural biases.

\section{Method}
\label{sec:method}
Let $G=(V,E)$ be a graph with feature matrix $X\in\mathbb{R}^{n\times d}$,
adjacency matrix $A$, and labels on a training-node index set
$\Train\subseteq\{1,\ldots,n\}$. Write $n_{\mathrm{tr}}=|\Train|$.
\SRC{} has five stages: graph signal construction, Fisher coordinate
selection, class subspace fitting, multi-alpha ridge fitting, and score fusion.

\begin{figure}[t]
    \centering
    \includegraphics[width=0.95\linewidth]{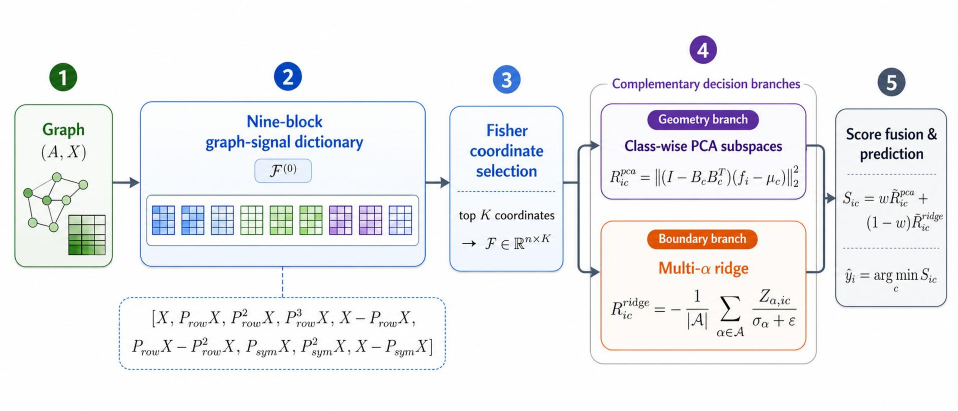}
    \caption{\SRC{} pipeline. The graph is converted into named signal blocks, discriminative coordinates are selected, and prediction is produced by fusing class-subspace residuals with a closed-form ridge boundary.}
    \label{fig:method}
\end{figure}

\subsection{Explicit multi-hop graph signal dictionary}
We use the row-normalized transition matrix $\Prw=D^{-1}A$ and the
symmetric-normalized matrix $\Psy=D^{-1/2}AD^{-1/2}$, following standard
normalized graph-operator conventions \citep{vonluxburg2007tutorial}. The
block-normalized
graph signal dictionary is
\begin{equation}
\Ffull
=
\bigl[
X,\;
\Prw X,\;
\Prw^2 X,\;
\Prw^3 X,\;
X-\Prw X,\;
\Prw X-\Prw^2 X,\;
\Psy X,\;
\Psy^2 X,\;
X-\Psy X
\bigr].
\label{eq:dictionary}
\end{equation}
Each block is row-$\ell_2$ normalized before concatenation, so
$\Ffull\in\mathbb{R}^{n\times p}$ with $p=9d$. After Fisher selection we
obtain a coordinate set $\Sel\subseteq\{1,\ldots,p\}$ with $|\Sel|=K$, and we
write
\begin{equation}
\Fsel = \Ffull_{[:,\Sel]}\in\mathbb{R}^{n\times K},
\qquad
\Ftr = \Fsel_{\Train,:}\in\mathbb{R}^{n_{\mathrm{tr}}\times K}.
\label{eq:selected-matrix}
\end{equation}
The downstream PCA, ridge, and atlas computations all use these selected
coordinates unless stated otherwise.

\subsection{Fisher coordinate selection}
For coordinate $j$ of the full dictionary $\Ffull$, the Fisher score is
\begin{equation}
q_j
=
\frac{\sum_{c=1}^C n_c(\mu_{c,j}-\mu_j)^2}
{\sum_{c=1}^C\sum_{i\in\Train:\,y_i=c}(\Ffull_{ij}-\mu_{c,j})^2+\epsilon}.
\label{eq:fisher}
\end{equation}
The numerator measures between-class separation and the denominator measures
within-class scatter. The top $K$ coordinates define the selected set
$\Sel$ in Eq.~\eqref{eq:selected-matrix}. The value of $K$ is selected by
validation.

\subsection{Class subspace residuals}
For each class $c$, PCA is fit on the selected training matrix $\Ftr$
restricted to class $c$. Let $\mu_c\in\mathbb{R}^{K}$ be the class center,
let $B_c\in\mathbb{R}^{K\times r_c}$ be the orthonormal basis selected by an
energy threshold, and let
$f_i=\Fsel_{i,:}^{\top}\in\mathbb{R}^{K}$ denote the selected feature vector
of node $i$. The class-subspace residual score is
\begin{equation}
R^{\mathrm{pca}}_{ic}
=
\left\|
(I-B_cB_c^\top)(f_i-\mu_c)
\right\|_2^2.
\label{eq:pcares}
\end{equation}
A low residual means that the node lies near the class geometry; a high
residual means that the node is poorly explained by that class subspace.
Unlike a black-box embedding, $r_c$, $\mu_c$, $B_c$, and
$R^{\mathrm{pca}}_{ic}$ can all be inspected directly.

\subsection{Closed-form multi-alpha ridge boundary}
PCA residuals capture class geometry, but class boundaries may still be better
described by a discriminative linear separator. We therefore fit a ridge
classifier in closed form. Let
$\Ytr\in\mathbb{R}^{n_{\mathrm{tr}}\times C}$ be the one-hot label matrix on
the training nodes and let $\Alphas$ denote the candidate regularization set.
For $\alpha\in\Alphas$,
\begin{equation}
\beta_\alpha
=
(\Ftr\Ftr^\top+\alpha I_{n_{\mathrm{tr}}})^{-1}\Ytr,
\qquad
Z_\alpha
=
\Fsel\Ftr^\top\beta_\alpha,
\label{eq:ridge}
\end{equation}
where $\beta_\alpha\in\mathbb{R}^{n_{\mathrm{tr}}\times C}$ and
$Z_\alpha\in\mathbb{R}^{n\times C}$. In the implementation, each score matrix
is first normalized by a single training-split standard deviation,
\begin{equation}
\sigma_\alpha
=
\operatorname{std}\bigl(\{(Z_\alpha)_{ic}: i\in\Train,\ c=1,\ldots,C\}\bigr),
\label{eq:ridge-std-alpha}
\end{equation}
and the residual-like ridge score is then defined by
\begin{equation}
R^{\mathrm{ridge}}_{ic}
=
-\frac{1}{|\Alphas|}
\sum_{\alpha\in\Alphas}
\frac{(Z_\alpha)_{ic}}{\sigma_\alpha+\epsilon}.
\label{eq:ridge-score}
\end{equation}
Smaller values are therefore better, matching the PCA-residual convention.

\subsection{Score fusion and prediction}
Before fusion, each branch is rescaled by a training-split standard deviation:
\begin{equation}
\sigma_{\mathrm{pca}}
=
\operatorname{std}\bigl(\{R^{\mathrm{pca}}_{ic}: i\in\Train,\ c=1,\ldots,C\}\bigr),
\qquad
\sigma_{\mathrm{ridge}}
=
\operatorname{std}\bigl(\{R^{\mathrm{ridge}}_{ic}: i\in\Train,\ c=1,\ldots,C\}\bigr).
\label{eq:branch-std}
\end{equation}
We then define
\begin{equation}
\widetilde R^{\mathrm{pca}}_{ic}
=
\frac{R^{\mathrm{pca}}_{ic}}{\sigma_{\mathrm{pca}}+\epsilon},
\qquad
\widetilde R^{\mathrm{ridge}}_{ic}
=
\frac{R^{\mathrm{ridge}}_{ic}}{\sigma_{\mathrm{ridge}}+\epsilon}.
\label{eq:branch-normalization}
\end{equation}
The final fused score is
\begin{equation}
S_{ic}
=
w\,\widetilde R^{\mathrm{pca}}_{ic}
+
(1-w)\,\widetilde R^{\mathrm{ridge}}_{ic},
\qquad
\hat y_i=\arg\min_c S_{ic}.
\label{eq:fusion}
\end{equation}
All hyperparameters, including $K$, the PCA dimension cap, the energy
threshold, the ridge-alpha set, and $w$, are selected by validation accuracy.
The algorithm therefore remains a validation-selected white-box classifier
rather than a trained neural network.

\subsection{Node-level signal atlas}
\label{sec:node-atlas-main}

The same scaffold used for prediction also produces the node-level atlas.
Fisher-selected coordinates retain their dictionary-block identities, so each
node can be decomposed over the explicit graph-signal dictionary. Let
$\Blocks$ denote the named dictionary blocks in Eq.~\eqref{eq:dictionary}, and
let $\BlockIdx_b$ be the coordinate indices belonging to block $b$ in the full
dictionary. For each selected block
\[
\Sel_b=\Sel\cap\BlockIdx_b,
\]
we define the Fisher-weighted block evidence of node $i$ as
\begin{equation}
E_b(i)=
\begin{cases}
\displaystyle
\frac{1}{|\Sel_b|}
\sum_{j\in\Sel_b}
|\Ffull_{ij}|\,q_j,
& |\Sel_b|>0,\\[1.0ex]
0, & |\Sel_b|=0.
\end{cases}
\label{eq:main-block-evidence}
\end{equation}
The corresponding block share is
\begin{equation}
\pi_b(i)
=
\frac{E_b(i)}
{\sum_{b'\in\Blocks}E_{b'}(i)+\epsilon}.
\label{eq:main-block-share}
\end{equation}
Thus, $\pi_b(i)$ is not a learned attention weight; it is a deterministic,
Fisher-weighted block readout induced by the fixed graph-signal dictionary and
the supervised Fisher selector.

Because the raw, low-pass, and high-pass families contain unequal numbers of
blocks, family-level composition is computed by first averaging evidence within
each family and then normalizing across families. For
$g\in\{\mathrm{raw},\mathrm{low},\mathrm{high}\}$, let $\Blocks_g$ be the
blocks in family $g$ and define
\begin{equation}
\bar E_g(i)
=
\frac{1}{|\Blocks_g|}
\sum_{b\in\Blocks_g}E_b(i),
\qquad
R_g(i)
=
\frac{\bar E_g(i)}
{\bar E_{\mathrm{raw}}(i)+\bar E_{\mathrm{low}}(i)+\bar E_{\mathrm{high}}(i)+\epsilon}.
\label{eq:main-family-adjusted-share}
\end{equation}
We write these three family-size-adjusted shares as
$\Rraw(i)$, $\Rlow(i)$, and $\Rhigh(i)$. They are the node-level signal
coordinates used in the graph-signal simplex and the dense phase portraits.
Let
\[
\mathcal{C}_{\Train}
=
\{\,c:\exists i\in\Train \text{ such that } y_i=c\,\}
\]
denote the set of classes present in the training split.

The atlas also records the branch-wise predictions

\begin{equation}
\hat y_i^{\mathrm{pca}}
=
\arg\min_c \widetilde R^{\mathrm{pca}}_{ic},
\qquad
\hat y_i^{\mathrm{ridge}}
=
\arg\min_c \widetilde R^{\mathrm{ridge}}_{ic},
\label{eq:main-branch-predictions}
\end{equation}
and the true-versus-nearest-wrong branch margins
\begin{equation}
M_i^{\mathrm{pca}}
=
\min_{c\neq y_i}\widetilde R^{\mathrm{pca}}_{ic}
-
\widetilde R^{\mathrm{pca}}_{i,y_i},
\qquad
M_i^{\mathrm{ridge}}
=
\min_{c\neq y_i}\widetilde R^{\mathrm{ridge}}_{ic}
-
\widetilde R^{\mathrm{ridge}}_{i,y_i}.
\label{eq:main-branch-margins}
\end{equation}
If the true label $y_i$ is absent from the training-class set
$\mathcal{C}_{\Train}$ for a particular split, these branch margins are treated
as undefined and are recorded as missing values in the implementation.

Together with the final prediction $\hat y_i$, final correctness, degree, and
the decision quadrant---both modules correct, PCA only, ridge only, or both
wrong---these records form a dense node-level atlas. Aggregating them gives
the dataset fingerprint used in Section~\ref{sec:mechanistic-atlas}: signal
composition, PCA--Ridge decision structure, class-subspace complexity, and
correct-versus-wrong signal shifts.

\section{Experimental Setup}
\textbf{Datasets.} We evaluate on six PyTorch Geometric node-classification
datasets \citep{fey2019pyg}: \dataset{Amazon-Computers},
\dataset{Amazon-Photo}, \dataset{Chameleon}, \dataset{Cornell},
\dataset{Texas}, and \dataset{Wisconsin}. The Amazon graphs are co-purchase
graphs from the benchmark studied by \citet{shchur2018pitfalls}. Chameleon is
loaded from \texttt{WikipediaNetwork} with
\texttt{geom\_gcn\_preprocess=True}, and the three WebKB graphs are loaded
from \texttt{WebKB}; these datasets are standard heterophily benchmarks
associated with the Geom-GCN setting \citep{pei2020geomgcn}, with Chameleon
originating from the Wikipedia networks of \citet{rozemberczki2021multi}.

\textbf{Baselines.} For the aligned main comparison, we rerun a disclosed
candidate pool under the same dataset-repeat protocol as \SRC{} and compare
against the strongest reproduced comparator for each dataset. This assignment
is used only to form the comparison target; it is not used to train, tune, or
select \SRC{}. The candidate pool includes GraphSAGE
\citep{hamilton2017graphsage}, GAT \citep{velickovic2018gat}, APPNP
\citep{gasteiger2019ppnp}, Correct and Smooth \citep{huang2021correct},
GCNII \citep{chen2020gcnii}, SIGN \citep{frasca2020sign}, LINKX
\citep{lim2021linkx}, and an MLP-propagation baseline. The strongest
reproduced baseline is GraphSAGE for all datasets except
\dataset{Chameleon}, where it is LINKX. All baseline epoch, model-state, and
optional hyperparameter choices use validation accuracy only; the test mask is
never used for training, early stopping, or selection.

\textbf{Evaluation.} We report mean accuracy and sample standard deviation
over ten repeated class-balanced random splits. All main predictive-validity
and white-box atlas figures use the six original evaluated datasets. The
fingerprint-conditioned black-box probing conditions are described separately
in Section~\ref{sec:atlas-diagnostic-probe} and
Appendix~\ref{app:prototype-construction}.

\textbf{Diagnostic protocol.} Predictor selection is separated from dataset
probing. The full \SRC{} scaffold is first selected by validation accuracy and
evaluated on the test set under the fixed selected configuration. The fitted
scaffold is then used as a graph-dataset probe whose atlas records
signal-family usage, class-subspace structure, branch behavior, margins, and
error locations. Correctness-dependent atlas quantities are computed only after
ordinary evaluation and are not used for training, hyperparameter selection,
model selection, or black-box architecture selection.

\Needspace{0.60\textheight}
\section{Predictive Validity of the White-Box Scaffold}

The atlas is meaningful only if it is produced by a classifier that captures
useful predictive structure. We therefore first test whether \SRC{} remains
competitive under the same repeated-split protocol as the aligned reproduced
baselines.

\begin{table}[H]
\centering
\small
\caption{
Mean-level companion to Figure~\ref{fig:paired-delta-main}. Values are mean test
accuracy $\pm$ sample standard deviation over ten aligned repeats; baselines are
the strongest aligned CPU reruns.
}
\label{tab:main-results}
\begin{tabular}{llccc}
\toprule
Dataset & Strongest baseline & Baseline ($n=10$) & \SRC{} ($n=10$) & Gain \\
\midrule
Amazon-Computers & GraphSAGE & 76.84 $\pm$ 2.28 & 78.71 $\pm$ 1.15 & +1.87 \\
Amazon-Photo & GraphSAGE & 88.37 $\pm$ 1.86 & 88.76 $\pm$ 1.35 & +0.39 \\
Chameleon & LINKX & 71.56 $\pm$ 1.49 & 72.48 $\pm$ 1.85 & +0.92 \\
Cornell & GraphSAGE & 72.43 $\pm$ 6.47 & 75.41 $\pm$ 7.26 & +2.97 \\
Texas & GraphSAGE & 84.32 $\pm$ 4.73 & 86.32 $\pm$ 4.08 & +1.99 \\
Wisconsin & GraphSAGE & 83.33 $\pm$ 5.25 & 84.31 $\pm$ 4.24 & +0.98 \\
\midrule
\textbf{Average} & -- & \textbf{79.48} & \textbf{81.00} & \textbf{+1.52} \\
\bottomrule
\end{tabular}
\end{table}
\FloatBarrier

\begin{figure}[H]
    \centering
    \includegraphics[width=0.92\linewidth]{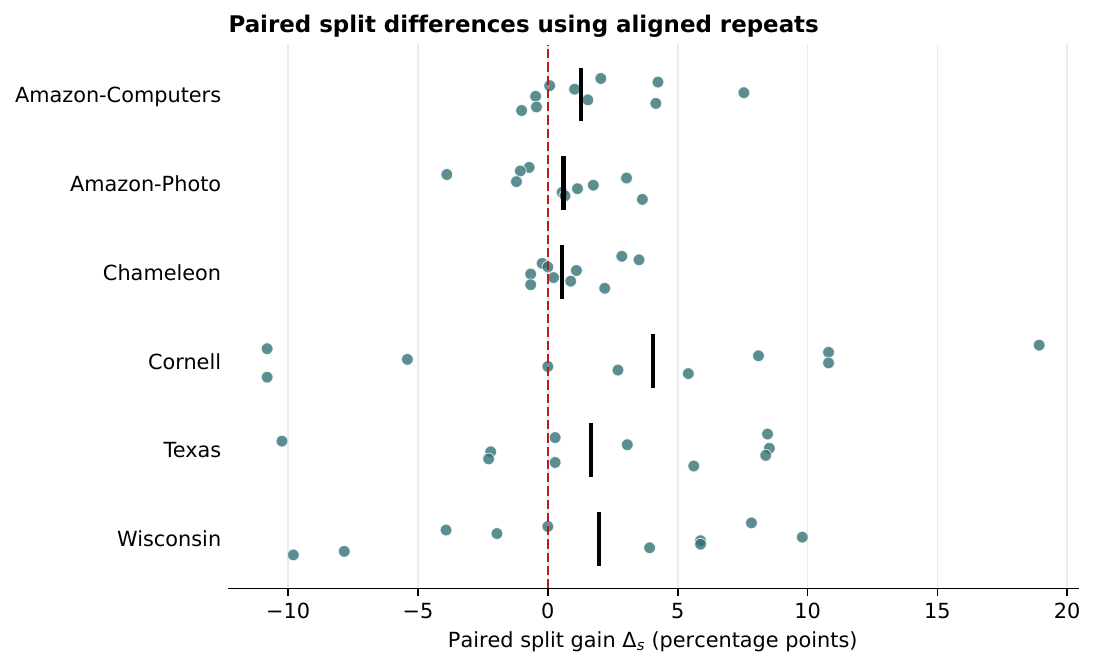}
    \caption{
    Split-level companion to Table~\ref{tab:main-results}. Each point is one
    matched dataset--repeat gain,
    $\Delta_s=\mathrm{Acc}^{(\mathrm{src})}_{s}-\mathrm{Acc}^{(\mathrm{base})}_{s}$,
    in percentage points; the dashed line marks zero.
    }
    \label{fig:paired-delta-main}
\end{figure}

\FloatBarrier

Table~\ref{tab:main-results} reports the absolute repeated-split comparison,
and Figure~\ref{fig:paired-delta-main} shows paired split-level differences
rather than only repeated means. \SRC{} achieves a positive mean gain on all
six datasets and a $+1.52$ percentage-point average gain over the strongest
aligned reproduced baseline. The split-level paired comparison shows that
\SRC{} wins most paired splits on five of six datasets and ties the win count
on \dataset{Wisconsin}; the full win-count summary is reported in
Appendix~\ref{app:paired-stability}. Treating the six dataset-level gains as
paired units gives a two-sided one-sample $t$-test $p=0.0105$, Wilcoxon
signed-rank $p=0.0313$, and sign test $p=0.0313$. These paired results support
the use of the fitted scaffold as a diagnostic probe: the subsequent atlas is
produced by a transparent model that preserves competitive predictive
performance under matched repeated splits.

\section{Mechanistic Interventions and White-box Mechanism Checks}
\label{sec:ablation-specialization}

We evaluate simplified \SRC{} variants under the same split, seed, and
train/validation/test protocol as the full scaffold. Each variant is selected
by validation accuracy only and evaluated over ten matched repeats. These
ablations serve as controlled interventions: by removing or isolating named
signal and decision components, they test whether structural changes agree
with the mechanisms measured by the atlas.

\Needspace{0.48\textheight}
\begin{table}[H]
\vspace{-0.5em}
\centering
\scriptsize
\caption{
Aligned mechanistic intervention summary. All variants are run with the same
ten split/seed protocol as the final full WG-SRC scaffold experiment. Entries are mean
test accuracy in percent. The bottom rows summarize cross-dataset
generalization. Rank is computed within each dataset among the listed variants;
lower is better. The full model is not always the best dataset-specific
specialist, but it has the best average rank, the best worst-case accuracy,
and is the only variant that remains in the top three on all six datasets. The
simplified variants are used as diagnostic interventions, not as a redefinition
of the main benchmark model.
}
\label{tab:ablation-specialization}
\resizebox{\linewidth}{!}{%
\begin{tabular}{lrrrrrrrl}
\toprule
Dataset
& \textbf{Full}
& Raw
& No high-pass
& No $P^3X$
& No sym
& PCA only
& Ridge only
& Best diagnostic variant \\
\midrule
Amazon-Computers & \textbf{78.71} & 63.46 & 79.73 & 76.32 & 73.54 & 76.83 & 78.58 & No high-pass (79.73) \\
Amazon-Photo & \textbf{88.76} & 76.88 & 90.27 & 87.16 & 84.78 & 87.15 & 89.12 & No high-pass (90.27) \\
Chameleon & \textbf{72.48} & 44.71 & 72.02 & 71.67 & 72.35 & 70.61 & 71.62 & Full / near-full (72.48) \\
Cornell & \textbf{75.41} & 77.30 & 71.08 & 75.14 & 74.32 & 66.76 & 77.57 & Ridge only (77.57) \\
Texas & \textbf{86.32} & 82.89 & 83.42 & 85.26 & 85.79 & 77.11 & 86.58 & Ridge only (86.58) \\
Wisconsin & \textbf{84.31} & 88.63 & 79.22 & 85.69 & 83.73 & 78.24 & 83.92 & Raw only (88.63) \\
\midrule
Mean accuracy & \textbf{81.00} & 72.31 & 79.29 & 80.21 & 79.09 & 76.12 & 81.23 & --- \\
Worst-case accuracy & \textbf{72.48} & 44.71 & 71.08 & 71.67 & 72.35 & 66.76 & 71.62 & --- \\
Average rank $\downarrow$ & \textbf{2.33} & 5.00 & 3.67 & 3.83 & 4.50 & 6.00 & 2.67 & --- \\
Top-3 count & \textbf{6/6} & 2/6 & 3/6 & 1/6 & 2/6 & 0/6 & 4/6 & --- \\
\bottomrule
\end{tabular}%
}
\end{table}
\FloatBarrier

Table~\ref{tab:ablation-specialization} shows that no single simplified
variant is uniformly reliable across datasets. Removing high-pass blocks
improves the two Amazon graphs, ridge-oriented variants are strong on Cornell
and Texas, and raw-only is strongest on Wisconsin. However, these specialized
gains do not transfer: raw-only collapses on Chameleon, PCA-only is weak across
the board, and no simplified variant appears in the top three on every
dataset.

The bottom rows of Table~\ref{tab:ablation-specialization} summarize this
generalist-versus-specialist pattern. The full scaffold has the best average
rank, the highest worst-case accuracy, and top-three performance on all six
datasets. Although Ridge-only attains a slightly higher mean accuracy, it
removes the class-subspace branch needed for class-subspace complexity,
PCA--Ridge complementarity, and the geometry-versus-boundary analysis in later
sections. We therefore keep the full scaffold as the main diagnostic base
model, while treating simplified variants as dataset-specific interventions.

These intervention results preview the atlas analysis. Amazon graphs behave as
low-pass-dominated regimes in which high-pass blocks can be removed for a
specialized improvement. Chameleon appears to benefit from the combined
multi-hop, high-pass, PCA, and ridge structure: the full scaffold is strongest,
while several near-full variants remain close. WebKB graphs are more boundary-
or raw-feature sensitive, making simplified specialists useful on particular
datasets. Thus, once a dataset has been measured, the same fingerprint can
suggest which mechanistic simplifications or emphases are worth testing first.

\section{From Node-Level Atlases to Dataset Fingerprints}
\label{sec:mechanistic-atlas}

The node-level atlas in Section~\ref{sec:node-atlas-main} records signal
shares, branch predictions, branch margins, correctness, degree, and
PCA--Ridge decision quadrants for each evaluated node. We call the aggregation
of these records an operational dataset fingerprint. It is operational because
it is computed from a fixed graph-signal dictionary and fixed white-box
decision modules; it is a dataset fingerprint because the same measurement
procedure yields comparable signal compositions, decision geometries, class
complexities, and error shifts across datasets. The fingerprint is not a claim
about the data-generating process itself. It measures how the fixed \SRC{}
scaffold uses a dataset.

For dataset $D$, let $\Eval_D$ denote the evaluation node set used for atlas
reporting; in the retrospective analysis below, this is the test-node set.
Using the training-class set $\mathcal{C}_{\Train}$ defined in
Section~\ref{sec:node-atlas-main}, write
$C_{\Train}=|\mathcal{C}_{\Train}|$. We define the dataset-level signal means by
\begin{equation}
R_D
=
\frac{1}{|\Eval_D|}
\sum_{i\in\Eval_D}\Rraw(i),
\qquad
L_D
=
\frac{1}{|\Eval_D|}
\sum_{i\in\Eval_D}\Rlow(i),
\qquad
H_D
=
\frac{1}{|\Eval_D|}
\sum_{i\in\Eval_D}\Rhigh(i),
\label{eq:dataset-signal-means}
\end{equation}
the mean class-subspace complexity by
\begin{equation}
C_D
=
\frac{1}{C_{\Train}}
\sum_{c\in\mathcal{C}_{\Train}} r_c,
\label{eq:dataset-complexity}
\end{equation}
and the PCA--Ridge decision fractions by
\begin{equation}
Q^{\mathrm{ridge}}_D
=
\frac{1}{|\Eval_D|}
\sum_{i\in\Eval_D}
\mathbf{1}\!\left\{
\hat y_i^{\mathrm{ridge}}=y_i,\;
\hat y_i^{\mathrm{pca}}\neq y_i
\right\},
\label{eq:ridge-only-fraction}
\end{equation}
\begin{equation}
Q^{\mathrm{hard}}_D
=
\frac{1}{|\Eval_D|}
\sum_{i\in\Eval_D}
\mathbf{1}\!\left\{
\hat y_i^{\mathrm{ridge}}\neq y_i,\;
\hat y_i^{\mathrm{pca}}\neq y_i
\right\}.
\label{eq:hard-fraction}
\end{equation}
Define the correct and wrong subsets of the evaluation nodes by
\begin{equation}
\Eval_D^{\mathrm{correct}}
=
\{\,i\in\Eval_D:\hat y_i=y_i\,\},
\qquad
\Eval_D^{\mathrm{wrong}}
=
\{\,i\in\Eval_D:\hat y_i\neq y_i\,\}.
\label{eq:correct-wrong-sets}
\end{equation}
For the correctness-dependent high-pass shift, let
\begin{equation}
H_D^{\mathrm{correct}}
=
\frac{1}{|\Eval_D^{\mathrm{correct}}|}
\sum_{i\in\Eval_D^{\mathrm{correct}}}\Rhigh(i),
\qquad
H_D^{\mathrm{wrong}}
=
\frac{1}{|\Eval_D^{\mathrm{wrong}}|}
\sum_{i\in\Eval_D^{\mathrm{wrong}}}\Rhigh(i),
\label{eq:correct-wrong-high}
\end{equation}
and define
\begin{equation}
\Delta H_D
=
H_D^{\mathrm{wrong}}
-
H_D^{\mathrm{correct}}.
\label{eq:highpass-shift}
\end{equation}
We then summarize the operational atlas of dataset $D$ by
\begin{equation}
\mathbf{m}(D)
=
\left[
R_D,\,
L_D,\,
H_D,\,
C_D,\,
Q^{\mathrm{ridge}}_D,\,
Q^{\mathrm{hard}}_D,\,
\Delta H_D
\right].
\label{eq:atlas-fingerprint}
\end{equation}

In the present paper, these quantities are reported on test nodes to
characterize how the fixed white-box probe behaves on the dataset after
standard supervised evaluation. The fingerprint contains two kinds of atlas
quantities. The signal-composition and class-geometry terms
$R_D,L_D,H_D,$ and $C_D$ are correctness-free summaries of how the fitted
scaffold uses the evaluated dataset. By contrast,
$Q^{\mathrm{ridge}}_D$, $Q^{\mathrm{hard}}_D$, and $\Delta H_D$ are
correctness-dependent audit quantities: they locate where the already
evaluated predictor succeeds or fails in the same atlas coordinates. These
quantities are used only as retrospective same-dataset diagnostic evidence,
not as inputs to training, hyperparameter selection, blind model choice, or
test-set-tuned redesign. If related diagnostic procedures are used
prospectively, correctness-dependent quantities should be computed only on
training or validation nodes.

\subsection{Graph-signal fingerprints}

\begin{table}[t]
\centering
\small
\caption{Family-size-adjusted node-level graph signal mixture. Values are average test-node signal-family shares in percent. Family composition is computed by first averaging Fisher-weighted block evidence within each signal family and then normalizing across the three families, thereby removing the trivial family-cardinality prior of the raw/low-pass/high-pass partition.}
\label{tab:signal-mix}
\begin{tabular}{lrrrr}
\toprule
Dataset & $n_{test}$ & Raw & Low-pass & High-pass \\
\midrule
Amazon-Computers & 13252 & 15.32 & 69.76 & 14.92 \\
Amazon-Photo & 7250 & 13.35 & 71.97 & 14.68 \\
Chameleon & 456 & 0.31 & 61.32 & 38.37 \\
Cornell & 37 & 28.12 & 34.99 & 36.89 \\
Texas & 38 & 29.47 & 30.59 & 39.94 \\
Wisconsin & 51 & 34.59 & 32.70 & 32.71 \\
\bottomrule
\end{tabular}
\end{table}

\begin{figure}[p]
\centering
\captionsetup[subfigure]{font=small,skip=1pt}
\captionsetup{font=small,skip=2pt}
\setlength{\abovecaptionskip}{2pt}
\setlength{\belowcaptionskip}{0pt}

\begin{subfigure}[t]{0.49\linewidth}
    \centering
    \includegraphics[width=\linewidth,height=0.285\textheight,keepaspectratio]{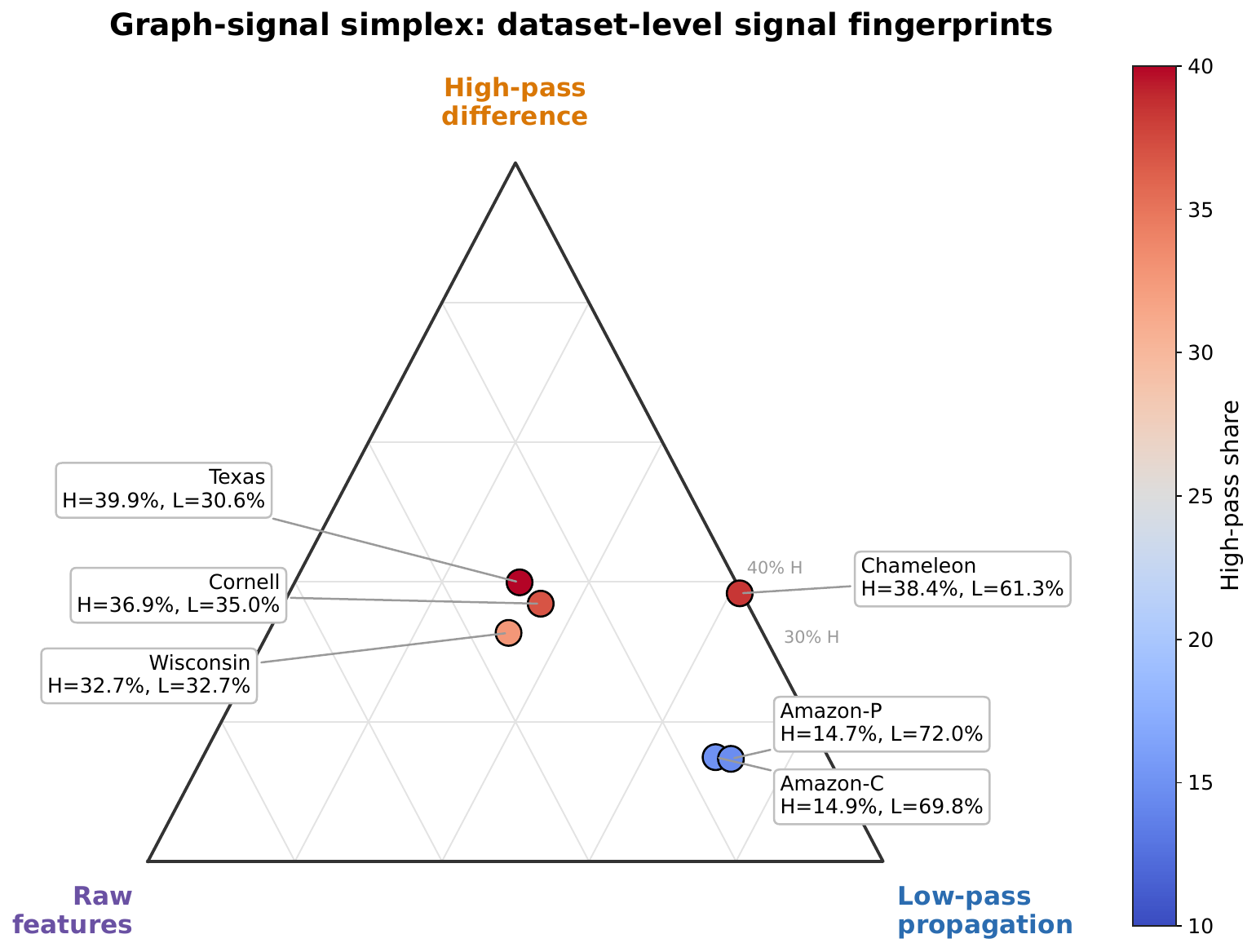}
    \caption{Dataset-level signal simplex.}
    \label{fig:atlas-grid-simplex}
\end{subfigure}
\hfill
\begin{subfigure}[t]{0.49\linewidth}
    \centering
    \includegraphics[width=\linewidth,height=0.285\textheight,keepaspectratio]{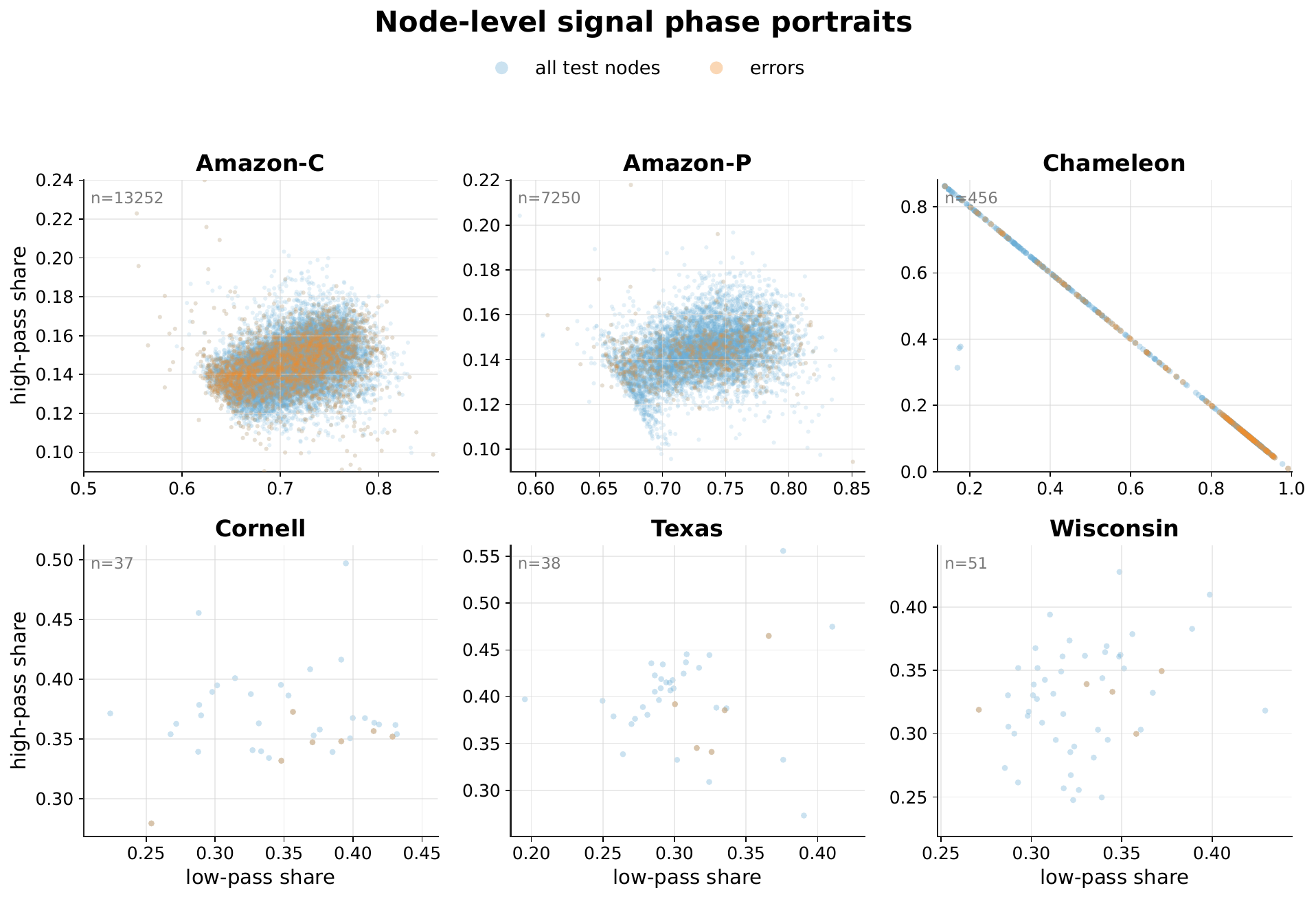}
    \caption{Node-level signal phase portraits.}
    \label{fig:atlas-grid-signal-phase}
\end{subfigure}

\vspace{0.55em}

\begin{subfigure}[t]{0.49\linewidth}
    \centering
    \includegraphics[width=\linewidth,height=0.285\textheight,keepaspectratio]{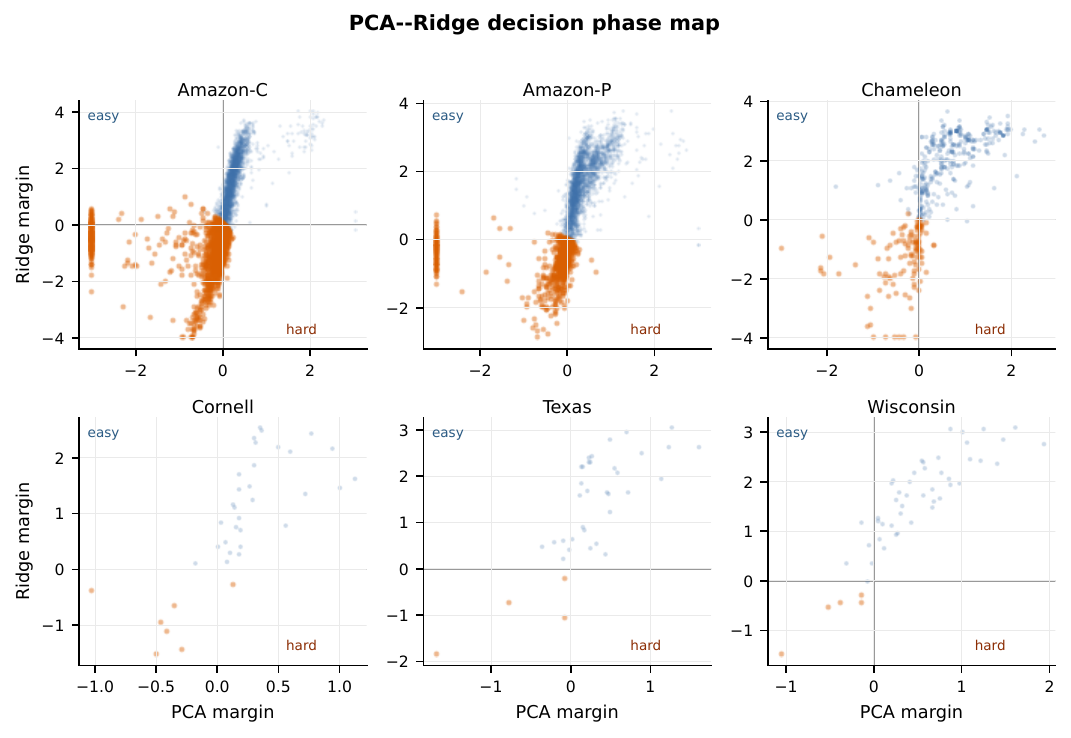}
    \caption{PCA--Ridge decision phase map.}
    \label{fig:atlas-grid-decision-phase}
\end{subfigure}
\hfill
\begin{subfigure}[t]{0.49\linewidth}
    \centering
    \includegraphics[width=\linewidth,height=0.285\textheight,keepaspectratio]{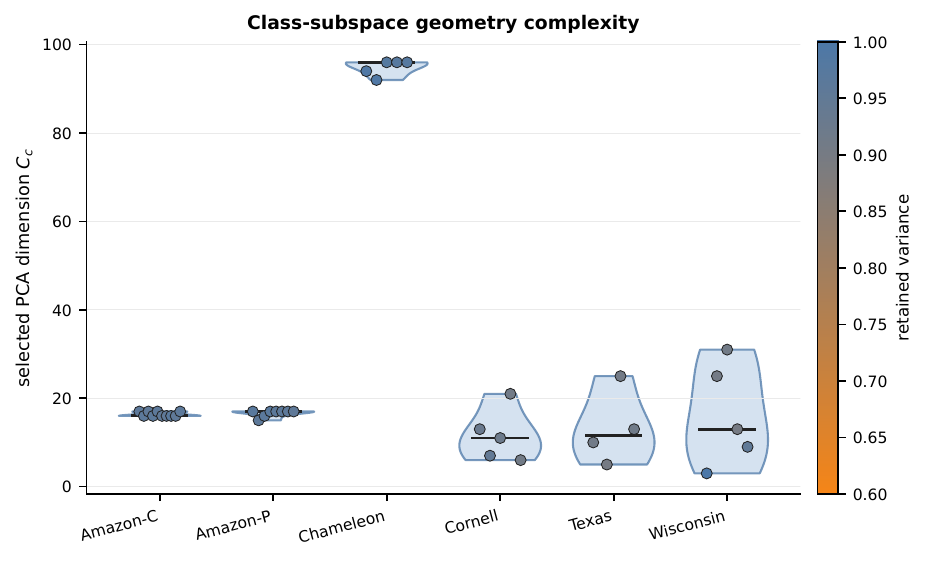}
    \caption{Class-subspace geometry complexity.}
    \label{fig:atlas-grid-class-geometry}
\end{subfigure}

\vspace{0.15em}

\caption{
Main-text atlas evidence for operational dataset fingerprints.
(a) The signal simplex summarizes family-size-adjusted raw, low-pass, and
high-pass signal composition.
(b) Node-level phase portraits localize errors inside the low-pass/high-pass
plane.
(c) PCA--Ridge phase maps separate geometry-driven, boundary-driven, easy, and
hard nodes.
(d) Class-subspace complexity shows the selected PCA dimension needed by each
class.
Together, the four panels connect dataset-level signal composition, node-level
error structure, decision-branch behavior, and class-subspace geometry.
}
\label{fig:atlas-main-grid}
\label{fig:atlas-signal-evidence}
\label{fig:atlas-decision-geometry}
\end{figure}

Table~\ref{tab:signal-mix} and Figure~\ref{fig:atlas-grid-simplex} report
family-size-adjusted signal-family shares, so the reported composition is not
mechanically driven by the unequal 1:5:3 block counts of the
raw/low-pass/high-pass partition. \dataset{Amazon-Computers} and
\dataset{Amazon-Photo} remain low-pass dominated, but at a corrected level of
roughly $70$--$72\%$ rather than the unadjusted $84$--$85\%$.
\dataset{Chameleon} shifts toward a mixed low-pass/high-pass profile, while
\dataset{Cornell}, \dataset{Texas}, and \dataset{Wisconsin} move into a more
raw- and high-pass-balanced regime. Thus, the classical
homophily/heterophily dichotomy can be refined into a measurable signal
composition: a graph may still be low-pass dominated, yet retain a nontrivial
raw or high-pass component after correcting for family cardinality.

Figures~\ref{fig:atlas-grid-signal-phase}--\ref{fig:atlas-grid-class-geometry}
provide the node- and class-level evidence behind the dataset fingerprint.
Figure~\ref{fig:atlas-grid-signal-phase} localizes errors inside the
low-pass/high-pass plane rather than only reporting dataset-level averages:
Amazon errors concentrate away from the dominant low-pass core, whereas
Chameleon occupies a broader mixed regime.
Figure~\ref{fig:atlas-grid-decision-phase} separates geometry-driven,
boundary-driven, easy, and hard nodes, showing why ridge-style boundary
correction is relevant on the WebKB graphs.
Figure~\ref{fig:atlas-grid-class-geometry} shows that Chameleon requires much
larger class subspaces than Amazon and WebKB, confirming that its selected
graph-signal geometry is substantially more complex. Together, these panels
close the main-text evidence chain from signal composition to node-level error
structure, decision-phase behavior, and class-subspace complexity.

\section{Fingerprint-Guided Diagnostic Guidance}
\label{sec:atlas-diagnostic-probe}

The mechanistic atlas turns a dataset from a single accuracy number into an
operational diagnostic object. Given the fingerprint $\mathbf{m}(D)$ in
Eq.~\eqref{eq:atlas-fingerprint}, \SRC{} asks which signal families,
class-subspace structures, and decision mechanisms are associated with success
or failure under the fitted white-box scaffold. The claim is not automatic
architecture search: the atlas provides post-evaluation, same-dataset
mechanism hypotheses that must be checked by aligned interventions and
validation-based model development. The operational claim is that the atlas can
prioritize follow-up analysis after evaluation: it identifies which mechanisms
should be suppressed, preserved, strengthened, or further tested first on the
same measured dataset. These mechanism hypotheses are then checked by aligned
interventions and by subsequent validation-based model development.

\begin{figure}[!htbp]
    \centering
    \includegraphics[
        width=0.86\linewidth,
        height=0.23\textheight,
        keepaspectratio
    ]{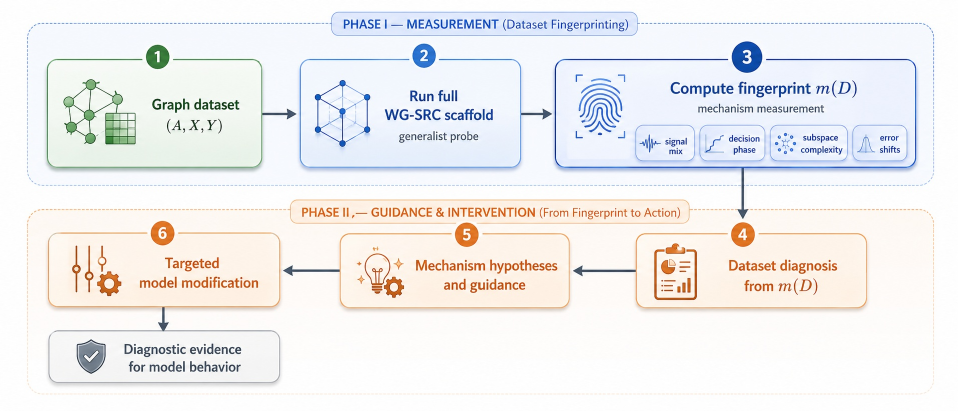}
    \vspace{-0.5em}
    \caption{
    Mechanism atlas as a dataset diagnostic probe. \SRC{} runs the full
    white-box scaffold, computes $\mathbf{m}(D)$, and uses it to generate
    post-evaluation, same-dataset mechanism hypotheses checked by aligned
    interventions and fingerprint-conditioned black-box probing.
    }
    \label{fig:atlas-guided-diagnosis-main}
\end{figure}

\begin{table}[!htbp]
\centering
\footnotesize
\caption{
From operational dataset fingerprint to same-dataset diagnostic guidance.
Each row maps an atlas signature to a mechanism diagnosis and a first
dataset-specific modification direction. The table is an operational
consistency map: the proposed directions are checked against aligned
white-box interventions and fingerprint-conditioned black-box component probes.
}
\label{tab:atlas-diagnostic-consistency}
\resizebox{\linewidth}{!}{%
\begin{tabular}{p{0.25\linewidth}p{0.27\linewidth}p{0.42\linewidth}}
\toprule
Atlas signature & Dataset diagnosis & Diagnostic guidance \\
\midrule
High $L_D$, low $H_D$, and positive $\Delta H_D$
& High-pass components may act as noise under the fitted scaffold
& Remove, downweight, gate, or regularize high-pass blocks \\
Near-zero $R_D$ and high $C_D$
& Raw-block evidence is weak and class geometry is complex
& Preserve graph-derived signals and improve class-specific subspace modeling \\
High $H_D$ and negative $\Delta H_D$
& Correct nodes show stronger high-pass evidence
& Preserve or adaptively gate high-pass differences rather than removing them globally \\
High $Q^{\mathrm{ridge}}_D$
& The ridge branch corrects some PCA-branch failures
& Strengthen the discriminative head or improve PCA--Ridge boundary fusion \\
High $Q^{\mathrm{hard}}_D$
& Both geometry and boundary branches fail on many nodes
& Investigate new signal blocks, hard-node treatment, uncertainty handling, or graph rewiring \\
High $R_D$ and low-to-moderate $C_D$
& Raw-block evidence is strong under the fitted scaffold
& Use raw-preserving skip design, weak propagation, or propagation gating \\
\bottomrule
\end{tabular}%
}
\end{table}

Table~\ref{tab:atlas-diagnostic-consistency} summarizes the diagnostic mapping
from atlas signatures to first modification directions. The supporting
white-box evidence comes from the aligned interventions in
Table~\ref{tab:ablation-specialization} and the correct-versus-wrong signal
shifts in Appendix~\ref{app:atlas-figures}. When the atlas suggests that a
mechanism is operationally harmful or necessary under the fitted scaffold, the
corresponding fixed intervention usually moves performance in the predicted
direction: Amazon benefits from removing high-pass blocks, Chameleon fails
under raw-only simplification, Texas shows strong boundary-oriented behavior,
and Wisconsin matches the raw-feature diagnosis. Thus, the fingerprint is not
only descriptive; it identifies which mechanisms are worth testing first for a
measured dataset.

Two measured conditions in the compact black-box summary are processed
prototypes used only for component probing. Amazon-Photo-LP+ strengthens a
low-pass regime by label-free mutual cosine-$k$NN edge densification, and
Wisconsin-R+ creates a raw-enhanced WebKB stress condition by label-free
rewiring/dropout processing. They are not used in the main predictive-validity
claim; they are controlled stress tests for asking whether sharper measured
dataset conditions expose corresponding black-box component biases.
Construction details are given in Appendix~\ref{app:prototype-construction}.

Table~\ref{tab:blackbox-compact-main} gives the corresponding minimum
black-box evidence. The full family-by-family component tables are reported in
Appendix~\ref{sec:blackbox-component-probing}, but the key pattern is already
visible in the compact summary: different measured fingerprints repeatedly
favor different architectural biases under fixed black-box component probes.

\begin{table}[!htbp]
\centering
\footnotesize
\caption{
Main-text compact summary of fingerprint-conditioned black-box component
probing. The table reports qualitative tendencies under fixed component
ablations across LINKX, GNNMoE, SGFormer, GOAT, NodeFormer, and tuned
GraphSAGE. Its purpose is to test whether measured dataset fingerprints
organize black-box component behavior, rather than to rank model families.
}
\label{tab:blackbox-compact-main}
\resizebox{\linewidth}{!}{%
\begin{tabular}{p{0.18\linewidth}p{0.27\linewidth}p{0.25\linewidth}p{0.23\linewidth}}
\toprule
Measured condition & Fingerprint diagnosis & Favored tendency & Usually disfavored tendency \\
\midrule
Amazon-Photo-LP+ &
Low-pass-dominant stress-test regime &
Shallow local propagation; original-graph or graph-branch behavior;
raw-preserving or no-high-pass designs &
Pure global attention; latent-only rewiring; unnecessary high-pass complexity \\
Chameleon &
Mixed low/high-pass, near-zero raw signal, high class-geometry complexity &
Graph-local structure; graph-branch dominance; selective structural bias &
Token-only, global-only, latent-only, or raw-dominant variants \\
Wisconsin-R+ &
Raw-enhanced WebKB regime with boundary sensitivity &
Root/raw-preserving shallow models; light original-graph use &
No-root ablations; deep structural complexity \\
Texas &
Boundary-sensitive and high-pass-balanced WebKB regime &
Shallow message passing; stronger discriminative or global mixing &
Over-fused variants; extra depth that dilutes the discriminative branch \\
Cornell &
Balanced WebKB control case &
Simple original-graph bias; shallow propagation; light or no fusion &
Deep heads, no-root ablations, or overly complex fusion modules \\
\bottomrule
\end{tabular}%
}
\end{table}
\FloatBarrier

Thus, the black-box probing evidence is not used to rank modern graph models.
Its role is narrower and diagnostic: once \SRC{} assigns an operational
fingerprint to a measured dataset, fixed component ablations show which
architectural directions are worth modifying first. This closes the main-text
evidence chain from white-box fingerprint, to aligned intervention, to
fingerprint-conditioned black-box behavior.

\section{Limitations}
\SRC{} trades computational simplicity for an explicit audit trail: graph-signal
construction, Fisher selection, class-wise PCA fitting, ridge solves, and
validation search all add cost, so Appendix~\ref{app:computational-profile}
reports runtime as a computational profile rather than a speed claim. The
method is also modular. Specialized variants can outperform the full scaffold
on some datasets, while the full model is kept as the analytically complete
base probe for producing the atlas. The baseline comparison is limited to the
disclosed aligned reproduced suite. Finally, the
fingerprint-conditioned black-box probing study is an initial same-dataset
reference-frame demonstration for dataset-specific modification: after a
dataset has been measured, the fingerprint prioritizes architectural directions
to test, while final model development still requires validation under the
target dataset protocol.

\section{Conclusion}
We presented \SRC, a white-box signal-subspace scaffold for graph
node classification and dataset diagnosis. Instead of learning hidden
message-passing representations, \SRC{} constructs explicit raw-feature,
low-pass, and high-pass graph-signal blocks, selects discriminative
coordinates, fits class-wise PCA subspaces, and combines them with a
closed-form ridge boundary. This gives the model an inspectable structure:
the same fitted variables used for prediction also define an operational
fingerprint of how a dataset is used by the scaffold.

The empirical results support this dual role. Under an aligned repeated-split
protocol on six datasets, \SRC{} remains competitive with selected reproduced
baselines while preserving the structure needed for diagnosis. The resulting
fingerprints separate low-pass-dominated Amazon graphs, mixed
high-pass and class-geometrically complex Chameleon behavior, and raw- or
boundary-sensitive WebKB graphs. Aligned white-box interventions agree with
these measurements: removing high-pass blocks helps when high-pass evidence is
noise-like, preserving graph-derived signals matters when raw features are
insufficient, and ridge-oriented decisions become important when class
geometry alone is inadequate.

The supported contribution is therefore operational rather than leaderboard
or causal in scope. After a standard validation-selected evaluation, \SRC{}
provides a reproducible dataset-under-scaffold fingerprint whose mechanism
hypotheses can guide same-dataset analysis and targeted model modification.
The black-box component probes in Appendix~\ref{sec:blackbox-component-probing}
extend this role by using the measured fingerprints as a reference frame for
testing which architectural biases are favored under different dataset
conditions. Future work can broaden this fingerprint-conditioned analysis to
larger benchmark suites, additional model families, and more controlled
dataset prototypes.

\appendix

\section*{Reproducibility Statement}
The package includes the LaTeX source, filtered summary tables, figures, CSV
files used to generate the tables and atlas figures, and the black-box
component-probing outputs used in the fingerprint-conditioned analyses.
All predictive-validity tables and white-box atlas figures use the six original
evaluated datasets. The fingerprint-conditioned black-box probing summaries
additionally include the processed prototype conditions described in
Appendix~\ref{app:prototype-construction}; these prototypes are not used in the
main predictive-validity claim. Model and baseline selections are
validation-only within each run, and paired split stability is computed by
matching dataset and repeat index.An anonymized supplementary source package contains the implementation scripts,
configuration files, split indices, random seeds, processed experiment outputs,
and figure-generation files needed to reproduce the reported tables and
figures.

\section*{Broader Impact Statement}
This work aims to make graph learning more transparent.
Potential positive impacts include easier auditing of graph models and better diagnosis of when graph propagation is harmful. Potential negative impacts are similar to those of node-classification systems generally: if applied to sensitive social, financial, or biological networks without care, predictions and explanations could still be misused. The proposed atlas should be treated as a diagnostic aid, not as a guarantee of fairness or causality.

\section{Paired Random-Split Stability}
\label{app:paired-stability}

The main text reports absolute repeated-split accuracy in
Table~\ref{tab:main-results} and visualizes split-level paired differences in
Figure~\ref{fig:paired-delta-main}. This appendix reports the paired
win-count table and keeps the formal paired-effect definitions and
significance-testing details.

\begin{table}[H]
\vspace{-0.5em}
\centering
\small
\caption{Paired random-split stability. For split $s$, $\Delta_s = Acc_{\mathrm{SRC},s}-Acc_{\mathrm{base},s}$ in percentage points, paired by dataset and repeat index. Baseline values are taken from the aligned CPU baseline rerun.}
\label{tab:stability-win}
\begin{tabular}{lrrrr}
\toprule
Dataset & \SRC{} wins & Mean $\Delta$ & Median $\Delta$ & Range $\Delta$ \\
\midrule
Amazon-Computers & 7/10 & +1.87 & +1.28 & [-1.01, +7.55] \\
Amazon-Photo & 6/10 & +0.39 & +0.61 & [-3.89, +3.64] \\
Chameleon & 7/10 & +0.92 & +0.55 & [-0.66, +3.51] \\
Cornell & 7/10 & +2.97 & +4.05 & [-10.81, +18.92] \\
Texas & 7/10 & +1.99 & +1.67 & [-10.24, +8.53] \\
Wisconsin & 5/10 & +0.98 & +1.96 & [-9.80, +9.80] \\
\bottomrule
\end{tabular}
\end{table}

\FloatBarrier

\subsection{Paired Significance Testing}
\label{sec:paired-significance}

Because \SRC{} and the strongest baseline are evaluated on matched
dataset--repeat pairs, the appropriate comparison is paired. For dataset $D$
and repeat $s$, we define the paired gain as

\begin{equation}
d_{D,s}
=
100\left(
\mathrm{Acc}^{(\mathrm{src})}_{D,s}
-
\mathrm{Acc}^{(\mathrm{base})}_{D,s}
\right).
\label{eq:paired-gain}
\end{equation}

where gains are measured in percentage points. Table~\ref{tab:paired-effect}
summarizes the paired effect size on each dataset without cluttering the main
text with six separate per-dataset hypothesis tests.

\begin{table}[t]
\centering
\footnotesize
\caption{
Paired effect summary against the strongest aligned baseline. Gains are
reported in percentage points. The effect size $d_z$ is paired Cohen's $d$,
computed as the mean paired gain divided by the standard deviation of the
split-level paired gains. The final row treats the six dataset-level mean
gains as the paired units.
}
\label{tab:paired-effect}
\begin{tabular}{lrrr}
\toprule
Dataset & Mean gain & 95\% CI & $d_z$ \\
\midrule
Amazon-Computers & $+1.87$ & $[-0.07,\,+3.81]$ & $0.69$ \\
Amazon-Photo     & $+0.39$ & $[-1.20,\,+1.98]$ & $0.18$ \\
Chameleon        & $+0.92$ & $[-0.14,\,+1.98]$ & $0.62$ \\
Cornell          & $+2.97$ & $[-4.05,\,+10.00]$ & $0.30$ \\
Texas            & $+1.99$ & $[-2.35,\,+6.33]$ & $0.33$ \\
Wisconsin        & $+0.98$ & $[-3.84,\,+5.81]$ & $0.15$ \\
\midrule
Dataset-level aggregate & $+1.52$ & $[+0.54,\,+2.50]$ & $1.62$ \\
\bottomrule
\end{tabular}
\end{table}

All six dataset-level mean gains are positive. To avoid treating the 60
split-level differences as fully independent observations, we report the
six dataset-level mean gains as the paired units for the aggregate stability
summary. A two-sided one-sample $t$-test over these six dataset-level gains
gives $p=0.0105$, a two-sided Wilcoxon signed-rank test gives $p=0.0313$, and
a two-sided sign test for all six gains being positive gives $p=0.0313$.
We use these tests as compact paired-protocol evidence rather than as a
standalone independent-dataset benchmark claim. Thus, the aggregate paired
evidence is consistent with positive dataset-level gains under the aligned
repeated-split protocol and supports the predictive validity of the fitted
scaffold as a diagnostic probe.

\section{Additional Atlas Views}
\label{app:atlas-figures}

The signal simplex, dense node-level signal phase portraits, PCA--Ridge
decision phase map, and class-subspace geometry complexity plot are included in
the main text as Figure~\ref{fig:atlas-main-grid}. This appendix retains only
the additional correct-versus-wrong signal-shift view, which provides a
complementary retrospective error diagnostic.

\subsection{Correct-vs-wrong signal shifts}

\begin{figure}[H]
    \centering
    \includegraphics[width=0.84\linewidth]{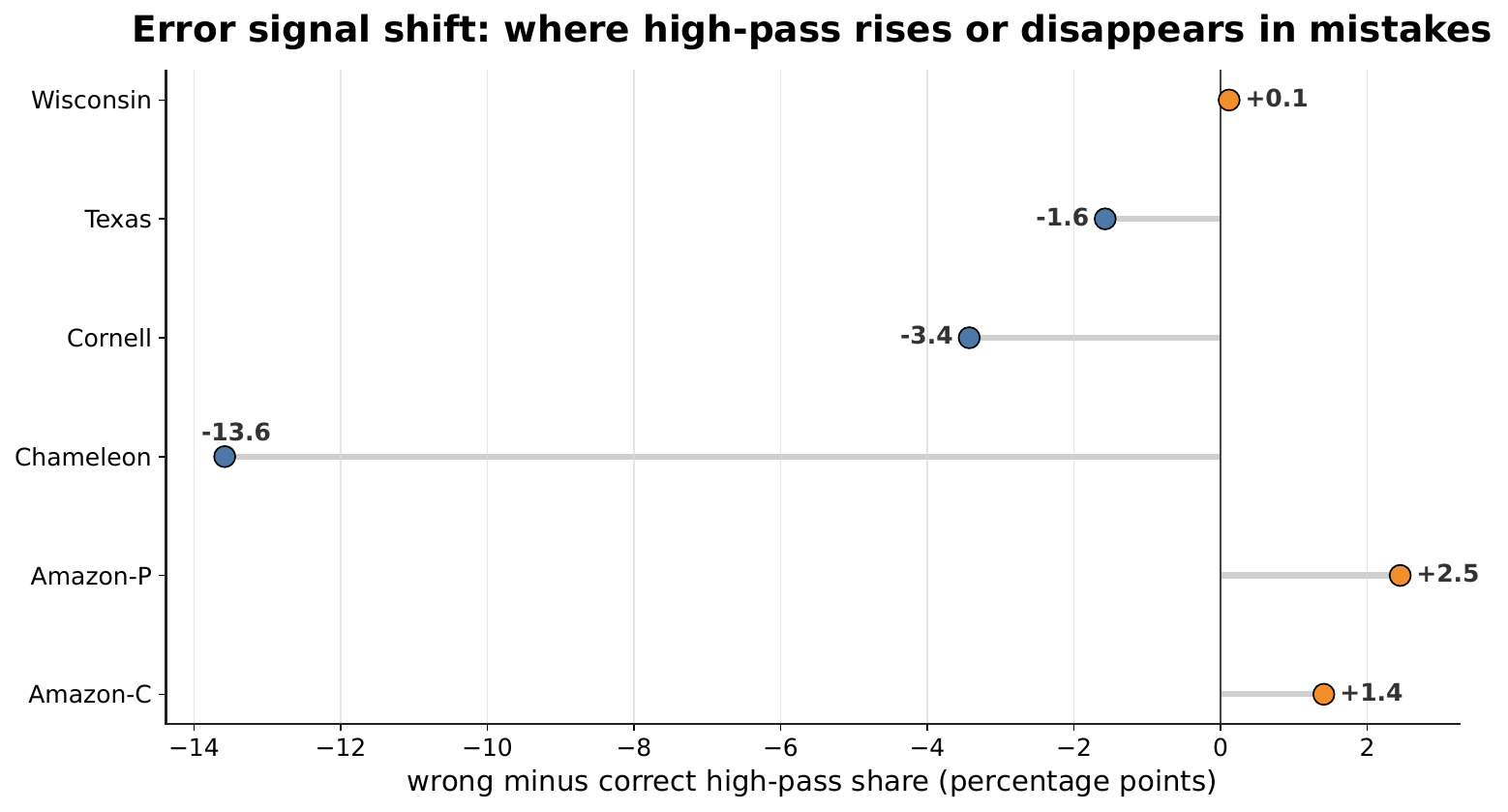}
    \caption{Error signal shift. The lollipop value is the family-size-adjusted high-pass share of wrong nodes minus the family-size-adjusted high-pass share of correct nodes. Positive values mean errors have more high-pass mass; negative values mean errors lose high-pass mass relative to correct nodes.}
    \label{fig:error-shift}
\end{figure}
\FloatBarrier

Figure~\ref{fig:error-shift} shows that errors are not uniformly distributed
in signal space. On Amazon graphs, errors carry more high-pass mass than
correct nodes, consistent with errors being atypical relative to the dominant
low-pass mechanism. On Chameleon, the shift goes in the opposite direction:
correct nodes rely more on high-pass differences, while errors are relatively
more low-pass. This directly supports the design choice of including
$X-\Prw X$ and $\Prw X-\Prw^2X$ in the dictionary.

\subsection{Fingerprint-conditioned black-box component probing}
\label{sec:blackbox-component-probing}

The aligned white-box interventions above establish that the atlas has
directional validity within the fitted \SRC{} scaffold. We now test whether
this role remains useful outside the \SRC{} scaffold itself. Once a dataset has
been measured by \SRC{}, we ask whether its operational fingerprint can serve
as a reference frame for black-box component probing: under the same
repeated-split protocol, do different measured dataset conditions repeatedly
favor different architectural biases? If so, then the fingerprint has
practical value beyond white-box interpretation alone, because it provides a
dataset-specific coordinate system for later same-dataset model analysis and
first-pass optimization guidance.

\paragraph{Processed prototype datasets.}
To sharpen the comparison, we used a small set of derived prototype datasets in
addition to the original graphs. These prototypes are \emph{not} introduced as
new benchmark claims and do not enter the paper's main predictive-validity
claim. Rather, they are label-free controlled stress-test environments used
only for fingerprint-conditioned black-box probing. Amazon-Photo-LP+ starts
from Amazon-Photo and adds mutual cosine-$k$NN edges in feature space
($k=10$), thereby strengthening a low-pass regime without using labels.
Wisconsin-R+ starts from Wisconsin and applies degree-preserving rewiring; if
that rewiring fails, the construction falls back to uniform edge dropout with
dropout fraction $0.15$. Chameleon, Texas, and Cornell are kept in their
original processed form as reference datasets. The purpose of these prototypes
is therefore not to manufacture higher scores or to redefine the benchmark, but
to provide controlled measured conditions under which black-box component
behavior can be compared more sharply than on the original graphs alone.

\paragraph{What is being tested.}
The earlier LINKX/GNNMoE study and the extended SGFormer, GOAT, NodeFormer, and
tuned GraphSAGE analysis all ask the same question: once a dataset fingerprint
has been measured, which \emph{kind} of black-box bias is favored by that
measured condition? This is a downstream use-case of the probe rather than a universal model
ranking. We fix the repeated-split protocol, use fixed within-family component
ablations, and ask whether different measured dataset conditions repeatedly
favor different architectural tendencies. Tables~\ref{tab:blackbox-by-family-fingerprint-a}
and~\ref{tab:blackbox-by-family-fingerprint-b} should therefore be read as a
fingerprint-conditioned probing map for dataset-specific model analysis rather
than as a universal ranking of black-box architectures.
For compactness, we abbreviate variant names inside the tables: nf-linear =
no\_fusion\_linear, no-hp-exp = no\_hp\_expert, g-only = graph\_only,
tok-only = token\_only, glob-only = global\_only, loc-only = local\_only,
s-head = shallow\_head, n-root = no\_root, and 1-layer = one\_layer.

\begin{table}[H]
\centering
\scriptsize
\setlength{\tabcolsep}{1.6pt}
\renewcommand{\arraystretch}{1.20}
\caption{
Fingerprint-conditioned black-box probing organized by model family
(part I). Each family entry reports the strongest variant and one informative
contrast on that measured dataset.
}
\label{tab:blackbox-by-family-fingerprint-a}
\begin{tabular}{P{0.12\linewidth}P{0.13\linewidth}P{0.21\linewidth}P{0.145\linewidth}P{0.145\linewidth}P{0.145\linewidth}}
\toprule
Dataset / role & Fingerprint type & Fingerprint summary & LINKX & GNNMoE & SGFormer \\
\midrule

Amazon-Photo-LP+ &
Low-pass-dominant processed prototype &
\makecell[tl]{$L$ dominant;\\high-pass error-associated} &
\bbcell{nf-linear (0.766)}{full (0.229)} &
\bbcell{no-hp-exp (0.698)}{full (0.261)} &
\bbcell{g-only (0.825)}{tok-only (0.210)} \\

Chameleon &
Mixed graph-structured heterophilic regime &
\makecell[tl]{Near-zero raw share;\\substantial high-pass;\\high class complexity} &
\bbcell{struct-only (0.636)}{x-only (0.468)} &
\bbcell{hard-top1 (0.598)}{full (0.555)} &
\bbcell{g-only (0.708)}{tok-only (0.223)} \\

Wisconsin-R+ &
Raw-enhanced processed prototype &
\makecell[tl]{Strong raw signal;\\some boundary sensitivity} &
\bbtie{nf-linear (0.842)}{x-only (0.842)} &
\bbcell{no-hp-exp (0.834)}{hard-top1 (0.697)} &
\bbcell{tok-only (0.739)}{g-only (0.456)} \\

Texas &
High-pass- and boundary-sensitive WebKB &
\makecell[tl]{High-pass active;\\clearer boundary correction} &
\bbcell{nf-linear (0.849)}{full (0.765)} &
\bbcell{no-hp-exp (0.868)}{full (0.838)} &
\bbcell{tok-only (0.703)}{g-only (0.565)} \\

Cornell &
Balanced WebKB control case &
\makecell[tl]{More balanced\\raw/high-pass;\\less extreme than Texas} &
\bbcell{x-only (0.773)}{full (0.735)} &
\bbtie{no-hp-exp (0.768)}{full (0.768)} &
\bbcell{tok-only (0.592)}{g-only (0.397)} \\

\bottomrule
\end{tabular}
\end{table}

\begin{table}[H]
\centering
\scriptsize
\setlength{\tabcolsep}{1.6pt}
\renewcommand{\arraystretch}{1.20}
\caption{
Fingerprint-conditioned black-box probing organized by model family
(part II). Each family entry reports the strongest variant and one informative
contrast on that measured dataset.
}
\label{tab:blackbox-by-family-fingerprint-b}
\begin{tabular}{P{0.12\linewidth}P{0.13\linewidth}P{0.21\linewidth}P{0.145\linewidth}P{0.145\linewidth}P{0.145\linewidth}}
\toprule
Dataset / role & Fingerprint type & Fingerprint summary & GOAT & NodeFormer & GraphSAGE \\
\midrule

Amazon-Photo-LP+ &
Low-pass-dominant processed prototype &
\makecell[tl]{$L$ dominant;\\high-pass error-associated} &
\bbnote{loc-only (0.261)}{family weak overall} &
\bbcell{g-only (0.757)}{latent-only (0.158)} &
\bbcell{s-head (0.896)}{n-root (0.882)} \\

Chameleon &
Mixed graph-structured heterophilic regime &
\makecell[tl]{Near-zero raw share;\\substantial high-pass;\\high class complexity} &
\bbcell{full (0.504)}{glob-only (0.223)} &
\bbcell{full (0.469)}{latent-only (0.229)} &
\bbcell{n-root (0.684)}{1-layer (0.613)} \\

Wisconsin-R+ &
Raw-enhanced processed prototype &
\makecell[tl]{Strong raw signal;\\some boundary sensitivity} &
\bbcell{glob-only (0.754)}{full (0.753)} &
\bbcell{g-only (0.828)}{latent-only (0.682)} &
\bbcell{1-layer (0.822)}{n-root (0.616)} \\

Texas &
High-pass- and boundary-sensitive WebKB &
\makecell[tl]{High-pass active;\\clearer boundary correction} &
\bbcell{glob-only (0.695)}{loc-only (0.624)} &
\bbcell{nf-linear (0.830)}{full (0.695)} &
\bbnote{1-layer (0.849)}{strongest overall} \\

Cornell &
Balanced WebKB control case &
\makecell[tl]{More balanced\\raw/high-pass;\\less extreme than Texas} &
\bbcell{s-head (0.592)}{full (0.514)} &
\bbcell{nf-linear (0.765)}{g-only (0.759)} &
\bbcell{1-layer (0.746)}{n-root (0.573)} \\

\bottomrule
\end{tabular}
\end{table}

\FloatBarrier

Several dataset-level patterns are now visible in a more structured way.
Amazon-Photo-LP+ is the cleanest low-pass prototype and consistently favors
shallow propagation and simple graph/original-signal use. Chameleon is the
most clearly graph-structured heterophilic regime: across families, useful
behavior repeatedly comes from graph-local structure rather than from raw-heavy,
pure-global, or latent-only variants. Wisconsin-R+ behaves as a raw-enhanced
prototype, where preserving ego/root information is more important than adding
architectural complexity. Texas is the clearest boundary-sensitive WebKB case:
some stronger global or discriminative mixing helps, but the strongest overall
behavior still comes from shallow message passing. Cornell serves as a more
balanced WebKB control, where simple original-graph and shallow-propagation
biases remain sufficient.

Taken together, these results show that the \SRC{} fingerprint is useful
beyond the \SRC{} scaffold itself. It does not merely interpret the fitted
white-box model; it also provides a practical reference frame for
dataset-specific black-box model optimization. Once a target dataset has been
assigned an operational fingerprint, fixed black-box component ablations reveal
which architectural biases are favored by that measured condition. A low-pass
fingerprint repeatedly favors shallow local propagation and no-high-pass
simplifications; a raw-enhanced fingerprint favors root-preserving designs; and
a boundary-sensitive fingerprint favors stronger discriminative or global
mixing. In this sense, the fingerprint acts as first-pass optimization
guidance: it does not choose the final model automatically, but it narrows the
initial set of architectural directions that a practitioner should test when
optimizing a model for the same measured dataset. This role is post-evaluation
and same-dataset in scope. The fingerprint prioritizes candidate modification
directions, while the final architecture, hyperparameters, and model state
remain selected by validation under the target dataset protocol.

\begin{table}[t]
\centering
\small
\caption{
Compact fingerprint-conditioned summary of black-box component tendencies.
This is a post-evaluation same-dataset summary, not a prospective model-selection
rule or an architecture-search table.
}
\label{tab:fingerprint-blackbox-summary}
\begin{tabular}{P{0.16\linewidth}P{0.24\linewidth}P{0.29\linewidth}P{0.23\linewidth}}
\toprule
Dataset & Fingerprint type & Favored bias & Usually disfavored \\
\midrule

Amazon-Photo-LP+ &
\makecell[tl]{Low-pass-dominant\\stress test} &
\makecell[tl]{Shallow local propagation;\\original-graph / graph-branch\\behavior; simple raw-preserving\\or no-high-pass designs} &
\makecell[tl]{Pure global attention;\\latent-only rewiring;\\deep heads; unnecessary\\high-pass complexity} \\

Chameleon &
\makecell[tl]{Mixed low/high-pass;\\near-zero raw; high\\class-geometry complexity} &
\makecell[tl]{Graph-local structure;\\graph-branch dominance;\\no-root aggregation;\\selective structural bias} &
\makecell[tl]{Token-only or global-only\\variants; latent-only rewiring;\\raw-dominant bias} \\

Wisconsin-R+ &
\makecell[tl]{Raw-enhanced\\WebKB regime} &
\makecell[tl]{Root/raw-preserving shallow\\models; one-layer message\\passing; light original-graph use} &
\makecell[tl]{No-root ablations;\\deep structural complexity;\\unnecessary latent/global fusion} \\

Texas &
\makecell[tl]{Boundary-sensitive,\\high-pass-balanced\\WebKB regime} &
\makecell[tl]{Shallow message passing;\\stronger discriminative/global\\mixing; original-graph branch} &
\makecell[tl]{Over-fused full variants;\\extra depth when it dilutes\\the discriminative branch} \\

Cornell &
\makecell[tl]{Balanced WebKB\\control case} &
\makecell[tl]{Simple original-graph bias;\\shallow propagation;\\light or no fusion} &
\makecell[tl]{Deep heads; no-root ablations;\\overly complex global/fusion\\modules} \\

\bottomrule
\end{tabular}
\end{table}

\section{Computational Profile}
\label{app:computational-profile}

Table~\ref{tab:efficiency} reports a computational profile for completeness.
It is not used as a main-text speed claim. Baseline runtime and accuracy are
taken from the aligned CPU baseline rerun; \SRC{} accuracy is standardized to
the ten-repeat main-table value. The purpose of this table is to document the
cost of the white-box audit trail: explicit graph-signal construction, Fisher
selection, class-wise PCA fitting, ridge solves, and validation search.

\begin{table}[H]
\centering
\small
\caption{
Appendix computational profile. Baseline runtime and accuracy are taken from
the aligned CPU baseline rerun; \SRC{} accuracy is standardized to the
ten-repeat main-table value. This table documents runtime cost and is not
intended as a speedup claim.
}
\label{tab:efficiency}
\begin{tabular}{llrrrr}
\toprule
Dataset & Best baseline & Baseline time(s) & \SRC{} time(s) & Ratio & \SRC{} acc. \\
\midrule
Amazon-Computers & GraphSAGE & 208.5 & 292.9 & 1.4$\times$ & 78.71 \\
Amazon-Photo & GraphSAGE & 81.3 & 136.4 & 1.7$\times$ & 88.76 \\
Chameleon & LINKX & 3.6 & 62.1 & 17.5$\times$ & 72.48 \\
Cornell & GraphSAGE & 0.8 & 4.5 & 5.7$\times$ & 75.41 \\
Texas & GraphSAGE & 0.8 & 4.1 & 5.2$\times$ & 86.32 \\
Wisconsin & GraphSAGE & 0.9 & 6.0 & 6.8$\times$ & 84.31 \\
\bottomrule
\end{tabular}
\end{table}

\section{Implementation Details}

\section{Implementation Details}
The Fisher coordinate budget was selected from
$K\in\{4000,5000,6000,8000\}$. For a dataset with dictionary dimension
$p=9d$, a candidate budget larger than $p$ was clipped to
\[
K_{\mathrm{eff}}=\min(K,p),
\]
so Fisher coordinate selection retained the top $K_{\mathrm{eff}}$
coordinates and retained all dictionary coordinates when $K\ge p$. The selected
coordinate set $\Sel$ in Eq.~\eqref{eq:selected-matrix} therefore has size
$|\Sel|=K_{\mathrm{eff}}$ in this corner case. The PCA maximum dimension was
selected from $r_{max}\in\{32,48,64,96\}$, with energy threshold
$\eta\in\{0.90,0.95,0.99\}$. Ridge regularizer sets were selected from
$\{0.01,0.1,1.0\}$, $\{0.05,0.5,5.0\}$, and $\{0.1,1.0,10.0\}$. The fusion
weight was selected from $w\in\{0.2,0.3,0.4,0.5,0.6,0.7,0.8\}$. All selections
were made using validation accuracy. The corresponding implementation and
experiment scripts are included in the anonymized supplementary package
described in the Reproducibility Statement.

\section{Class-Pair Geometry and Confusion}
\begin{figure}[H]
    \centering
    \includegraphics[width=0.82\linewidth]{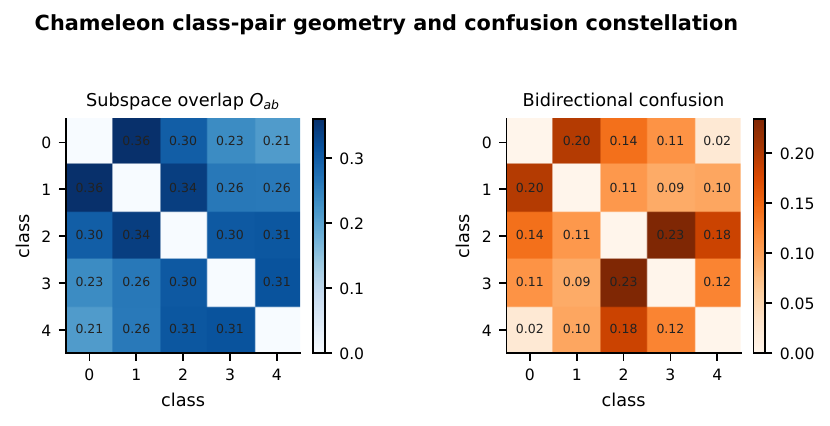}
    \caption{Chameleon class-pair geometry and confusion constellation. The left panel shows pairwise PCA subspace overlap; the right panel shows bidirectional confusion. This appendix figure illustrates how the same subspace objects used for prediction can also diagnose class-pair errors.}
    \label{fig:subspace-confusion}
\end{figure}

\section{Processed Prototype Construction and Scope}
\label{app:prototype-construction}

Table~\ref{tab:prototype-construction} documents the processed prototype
datasets used only in the fingerprint-conditioned black-box probing analysis.
These datasets are not part of the main predictive-validity claim and are not
introduced as new benchmark datasets. Their role is only to provide label-free
controlled stress-test environments under which black-box component behavior can
be compared under clearer measured dataset conditions.

\begin{table}[t]
\centering
\footnotesize
\caption{
Construction details of the processed prototype datasets used in the black-box
probing analysis. Edge counts are undirected edge counts after processing.
}
\label{tab:prototype-construction}
\begin{tabular}{P{0.17\linewidth}P{0.23\linewidth}P{0.20\linewidth}P{0.16\linewidth}P{0.18\linewidth}}
\toprule
Dataset & Construction & Key parameters & Processed size & Scope \\
\midrule
Amazon-Photo-LP+ &
Amazon-Photo plus mutual cosine-$k$NN edge densification in feature space &
$k=10$; label-free &
$7650$ nodes, $119250$ undirected edges &
Used only as a low-pass stress-test prototype in black-box probing \\

Wisconsin-R+ &
Wisconsin plus degree-preserving rewiring; if rewiring fails, use uniform edge-dropout fallback &
rewire fraction $=0.20$; fallback dropout fraction $=0.15$; label-free &
$249$ nodes, $449$ undirected edges &
Used only as a raw-enhanced stress-test prototype in black-box probing \\

Chameleon &
Original processed dataset retained as reference &
none &
$2277$ nodes, $31396$ undirected edges &
Reference dataset in black-box probing \\

Texas &
Original processed dataset retained as reference &
none &
$183$ nodes, $287$ undirected edges &
Reference dataset in black-box probing \\

Cornell &
Original processed dataset retained as reference &
none &
$183$ nodes, $278$ undirected edges &
Reference dataset in black-box probing \\
\bottomrule
\end{tabular}
\end{table}

\begin{table}[t]
\centering
\footnotesize
\caption{
Black-box families and fixed component ablations used in the
fingerprint-conditioned probing analysis. The purpose is not exhaustive
cross-family benchmarking, but controlled probing of different architectural
biases under measured dataset fingerprints.
}
\label{tab:blackbox-family-ablation-scope}
\begin{tabular}{P{0.16\linewidth}P{0.34\linewidth}P{0.38\linewidth}}
\toprule
Family & Fixed variants & Main architectural bias being probed \\
\midrule
LINKX &
full, x-only, struct-only, nf-linear, deep-head, deep-struct &
Raw/ego preservation versus structure branch, necessity of fusion, and depth placement \\

GNNMoE &
full, hard-top1, uniform-gate, no-hp-exp, deeper-experts, top2-gate &
Routing hardness, adaptive gating, explicit high-pass expert contribution, and expert depth \\

SGFormer &
full, tok-only, g-only, nf-linear, s-head, deep-head &
Global token mixing versus graph branch, fusion strength, and head depth \\

GOAT &
full, glob-only, loc-only, nf-linear, s-head, deep-head &
Global attention versus local propagation, fusion strength, and head depth \\

NodeFormer &
full, g-only, latent-only, nf-linear, s-head, deep-head &
Original-graph use versus learned latent graph structure, fusion strength, and head depth \\

GraphSAGE &
full, 1-layer, three-layer, n-root, s-head, deep-head &
Shallow versus deeper message passing, root/ego preservation, and head depth \\
\bottomrule
\end{tabular}
\end{table}

\end{document}